\documentclass[acmsmall]{acmart}
\AtBeginDocument{%
  \providecommand\BibTeX{{%
    \normalfont B\kern-0.5em{\scshape i\kern-0.25em b}\kern-0.8em\TeX}}}

\setcopyright{acmcopyright}
\copyrightyear{2018}
\acmYear{2018}
\acmDOI{XXXXXXX.XXXXXXX}

\usepackage{graphicx}
\usepackage{booktabs}
\usepackage{url}
\usepackage{color}
\usepackage{multirow}
\usepackage{amsmath}
\usepackage{amsthm}

\usepackage{amssymb}
\usepackage{caption}
\usepackage{subcaption}
\usepackage{xcolor}
\newtheorem{myDef}{Hypothesis}

\begin{document}

\title{Graph Contrastive Topic Model}

\author{Zheheng Luo}
\email{zheheng.luo@manchester.ac.uk}
\affiliation{%
  \institution{The University of Manchester}
  \streetaddress{Oxford Road}
  \city{Manchester}
  \country{UK}
  \postcode{M13 9PL}}
  
\author{Lei Liu}
\email{liulei95@whu.edu.cn}
\affiliation{%
  \institution{Wuhan University}
  \streetaddress{Wuchang District}
  \city{Wuhan}
  \state{Hubei}
  \country{China}
  \postcode{430072}
}

\author{Qianqian xie}
\authornote{corresponding author}
\email{xqq.sincere@gmail}
\author{Sophia Ananiadou}
\email{sophia.ananiadou@manchester.ac.uk}
\affiliation{%
  \institution{The University of Manchester}
  \streetaddress{Oxford Road}
  \city{Manchester}
  \country{UK}
  \postcode{M13 9PL}}

\renewcommand{\shortauthors}{Luo and Xie, et al.}

\begin{abstract}
Contrastive learning has recently been introduced into neural topic models to improve latent semantic discovery.
However, existing NTMs with contrastive learning suffer from the sample bias problem owing to the word frequency-based sampling strategy, which may result in false negative samples with similar semantics to the prototypes.
In this paper, we aim to explore the efficient sampling strategy and contrastive learning in NTMs to address the aforementioned issue.
We propose a new sampling assumption that negative samples should contain words that are semantically irrelevant to the prototype. 
Based on it, we propose the graph contrastive topic model (GCTM), which conducts graph contrastive learning (GCL) using informative positive and negative samples that are generated by the graph-based sampling strategy leveraging in-depth correlation and irrelevance among documents and words.
In GCTM, we first model the input document as the document word bipartite graph (DWBG), and construct positive and negative word co-occurrence graphs (WCGs), encoded by graph neural networks, to express in-depth semantic correlation and irrelevance among words.
Based on the DWBG and WCGs, we design the document-word information propagation (DWIP) process to perform the edge perturbation of DWBG, based on multi-hop correlations/irrelevance among documents and words.
This yields the desired negative and positive samples, which will be utilized for GCL together with the prototypes to improve learning document topic representations and latent topics. 
We further show that GCL can be interpreted as the structured variational graph auto-encoder which maximizes the mutual information of latent topic representations of different perspectives on DWBG.
Experiments on several benchmark datasets \footnote{\url{https://github.com/zhehengluoK/GCTM}} demonstrate  
the effectiveness of our method for topic coherence and document representation learning compared with existing SOTA methods.
\end{abstract}




\maketitle

\section{Introduction}
Recently, an increasing body of research~\cite{JMLR:v22:21-0089,NEURIPS2021_CLNTM} on neural topic models recognizes the efficacy of the contrastive learning~\cite{nips/ChuangRL0J20} in improving latent topic modelling.
Inspired by the success of contrastive learning in other areas~\cite{Oord/abs-1807-03748,gao-etal-2021-simcse}, existing attempts~\cite{JMLR:v22:21-0089,NEURIPS2021_CLNTM,wu2022mitigating} design the contrastive loss to guide the topic learning, with positive and negative samples by the word frequency based sampling strategy.
~\citet{JMLR:v22:21-0089} randomly split a document into two to form positive pairs and takes subsamples from two randomly chosen documents as negative samples.
~\citet{NEURIPS2021_CLNTM} generate positive/negative samples by replacing low-frequency words/high-frequency words in a given document.
For the short text, ~\citet{wu2022mitigating} find the positive and negative samples by the topic semantic similarity between two short text pairs.
Since they learn the topic semantics of the input document from its document word feature, their sampling strategy is essentially similar to that in~\citet{NEURIPS2021_CLNTM}. 
By learning to distinguish between positive and negative samples,
they are able to generate superior latent topics when compared to widely-used neural topic models (NTMs)~\cite{icml/MiaoYB16,iclr/SrivastavaS17-ProdLDA,gao2019incorporating,peng2018neural,peng2018bayesian,xie2021neural}.

\begin{figure}[!htb]
    \centering
    \includegraphics[width=0.8\textwidth]{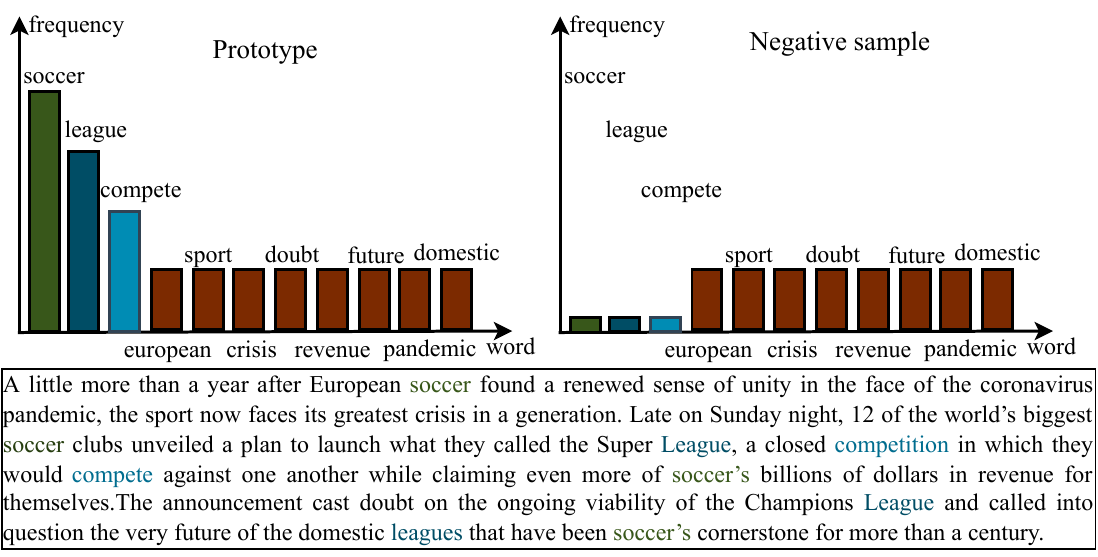}
    \caption{Example of the sample bias problem in CLNTM~\cite{NEURIPS2021_CLNTM}.}
    \label{fig:bias}
\end{figure}

However, their hypothesis on sampling negative samples that the ideal negative sample should exclude the high-frequency words of the input document as much as possible can be invalid and lead to the sample bias problem~\cite{nips/ChuangRL0J20,iclr/RobinsonCSJ21}.
Their negative samples can be false samples with similar semantics to the source document.
To highlight this phenomenon, we provide in Table~\ref{similar} the averaged similarity between the prototype and its negative sample in the SOTA neural topic model with the contrastive learning CLNTM~\cite{NEURIPS2021_CLNTM} on three benchmark datasets.
In Table~\ref{similar}, 
\begin{table}[!htb]
\caption{Comparison of the averaging similarities between the prototypes with their negative samples generated by our method and CLNTM~\cite{NEURIPS2021_CLNTM}, on benchmark datasets.}
\centering
\begin{tabular}{cccc}
\hline
\textbf{Dataset} & \textbf{20NG} & \textbf{IMDB} & \textbf{NIPS}\\
\hline
CLNTM  & 0.954 & 0.935 & 0.962\\ 
GCTM& 0.108 & 0.108 & 0.109\\ 
\hline
\end{tabular}
\label{similar}
\end{table}
it is discovered that the average similarity between the prototype and its negative sample generated by CLNTM is significantly high.
For each input document with the TF-IDF input feature $x$, CLNTM considers the top-k words with the highest TF-IDF score to be the main contributor of the topic for the input document and replaces TF-IDF scores of top-k words with the corresponding score of the reconstructed feature $\hat{x}$ by a neural topic model to generate negative samples $x^-$.
However, the topic of a document is not always determined by its high-frequency words, but also by other salient words~\cite{griffiths2019probabilistic,lau2014machine,chi2019topic}.
For instance, as shown in Figure \ref{fig:bias}, 
the input document describes the crisis in European soccer. 
Its negative sample still conveys a similar topic about a crisis in European sport as the prototype,
even though high-frequency words like "league", "soccer" and "compete" are removed.
This sample bias issue will mislead the model to shift the representations of the source document away from semantically identical false negatives, to impact the performance.

In this paper, we aim to explore sampling instructive negative and positive samples in the neural topic model scenario to address the sample bias issue.
Motivated by the new assumption that: \textit{the most beneficial negative samples should encompass as many distinct words as feasible that are semantically uncorrelated with the prototype}, we propose the graph-based sampling strategy guided by the in-depth correlation and irrelevance information among documents and words.
Based on this, we propose the novel graph contrastive neural topic model (GCTM), that 
models document data augmentation as the graph data augmentation problem and conduct graph contrastive learning (GCL) based on instructive positive and negative samples generated by the graph-based sampling strategy.
We build positive and negative word co-occurrence graphs (WCGs), and encode them with the graph neural networks (GNNs)~\cite{DBLP:conf/iclr/KipfW17} to capture multi-hop semantic correlation and irrelevance among words. 
The input document is also modelled as the document-word bipartite graph (DWBG) structured data.
Based on the DWBG and WCGs, we design the document-word information propagation (DWIP) process to perform the graph augmentation: edge perturbation on the DWBG.
It is able to identify new words that directly/distantly correlate to the input document to generate positive samples and recognizes words that are totally irrelevant to the input document to generate negative samples, by information propagation based on DWBG and WCGs.
This yields the desirable negative samples with words that are semantically distinct from prototypes, as well as positive samples with words that correlate to prototypes. 
As shown in Table ~\ref{similar}, the average similarity between the prototype and its negative sample generated by our method is significantly lower than that of CLNTM.
Moreover, we show that the GCTM with graph contrastive loss can be interpreted as the structured variational graph auto-encoder (VGAE)~\cite{kipf2016variational},
which explains its superiority in modelling topic posterior over previous NTMs that are variational auto-encoders (VAEs) with a single latent variable. 
The main contributions of our work are as follows:
\begin{enumerate}
    \item We propose GCTM, a new graph contrastive neural topic model 
    that models the contrastive learning problem of document modelling as the graph contrastive learning problem, to better capture the latent semantic structure of documents.
    \item We propose a novel graph-based sampling strategy for NTM based on graph data augmentation with the multi-hop semantic correlations and irrelevances among documents and words, that can generate more instructive positive and negative samples to enhance the effectiveness of the topic learning.
    \item Extensive experiments on three real-world datasets reveal that our method is superior to previous state-of-the-art methods for topic coherence and document representation learning.  
\end{enumerate}

\section{Related work}

\subsection{Contrastive Learning for Neural Topic Models}
\begin{figure*}[!hbt]
    \centering
    \includegraphics[width=\textwidth]{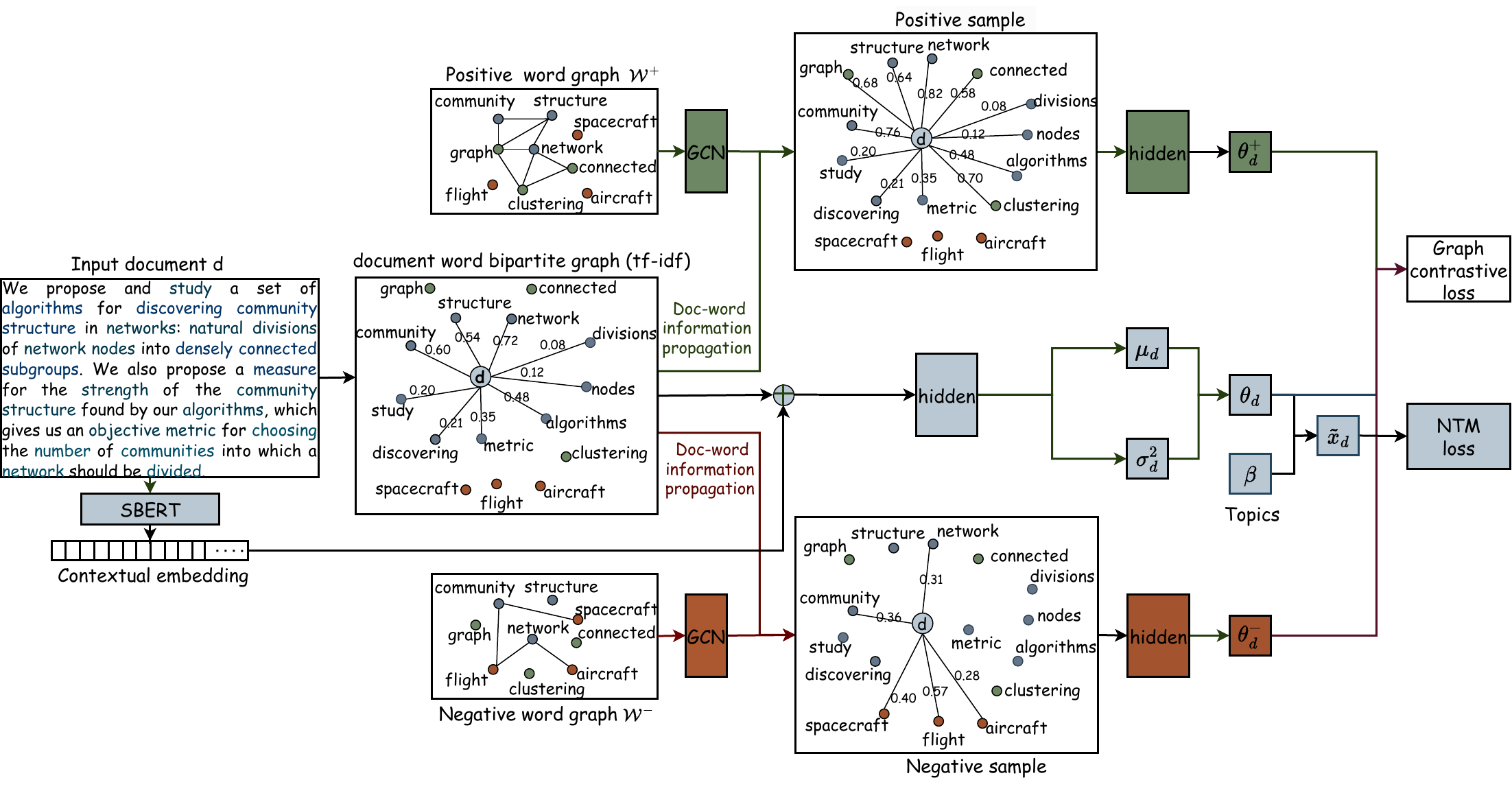}
    \caption{The architecture of GCTM.}
    \label{fig:my_label}
\end{figure*}
Recent research has incorporated contrastive learning into NTMs, motivated by the success of contrastive learning in many NLP tasks~\cite{2020-CERT,gao-etal-2021-simcse,luo2023citationsum}.
Contrastive learning is based on the concept of bringing similar pairs together and pushing dissimilar pairs apart in the latent space. 
Designing effective positive and negative samples is a vital step in contrastive learning,
especially for topic modeling, where even the substitution of a single word would change the whole sentence's meaning.
\citet{JMLR:v22:21-0089} provided theoretical insight for contrastive learning in a topic modeling setting.
They used paragraphs from the same text as the positive pairs, and paragraphs from two randomly sampled texts as the negative pairs.
However, their sampled negative pairs can be invalid and even introduce noise without considering semantic similarity when sampling negatives.
There is the possibility that paragraphs from two randomly selected texts may still share similar topics.
\citet{NEURIPS2021_CLNTM} proposed an approach to draw positive/negative samples by substituting $k$ tokens with the lowest/highest TF-IDF with corresponding reconstructed representation.
To tackle the data sparsity issue in short text topic modelling, \citet{wu2022mitigating} proposed to find the positive and negative samples based on topic semantic similarity between short texts, and then conduct the contrastive learning on the topic representations of input short texts.
They learn topic semantics from the document word features, therefore have a similar sampling strategy with \citet{NEURIPS2021_CLNTM}.
However, as previously mentioned, their sampling method may suffer from the sample bias problem due to their word frequency-based sampling strategy, which generates noisy negatives that are semantically related to the source document and misleads the model.
\subsection{Graph Neural Topic Models} 
In recent years neural topic models (NTMs) based on VAE~\cite{corr/KingmaW13} have received a surge of interest due to the flexibility of the Auto-Encoding Variational Bayes (AEVB) inference algorithm.
A number of NTMs emerge such as NVDM~\cite{icml/MiaoYB16}, ProdLDA \cite{iclr/SrivastavaS17-ProdLDA} and SCHOLAR~\cite{card-etal-2018-SCHOLAR} et al.
Recently, graph neural networks (GNNs) have been extensively used in NTMs, due to their success in modelling graph structure.
GraphBTM~\cite{zhu-etal-2018-GraphBTM} was the first attempt to encode the biterm graph using GCN~\cite{iclr/Kipf2017-GCN} to enhance the topic extraction quality of the biterm topic model (BTM, \citealp{www/Yan13-btm}) on short texts.
However, it was incapable of generating document topic representations and capturing semantic correlations between documents.
The following works used GNNs to encode the global document-word graph and the word co-occurrence graph, including GTM~\cite{zhou-etal-2020-GTM}, GATON~\cite{www/yang20-gaton}, GTNN~\cite{www/XieHDPN21}, DWGTM~\cite{wang-etal-2021-DWGTM}, and SNTM~\cite{bahrainian-etal-2021-SNTM}.
In addition to the bag-of-words (BoW), GNTM~\cite{NEURIPS2021_GNTM} considered the semantic dependence between words by constructing the directed word dependency graph.
However, no previous efforts have employed a contrastive framework that can improve the discovery of latent semantic patterns of documents, via data augmentation and optimizing the mutual information among the prototype, negative and positive samples.
In contrast to these methods, we aim to investigate the impact of effective data augmentation and contrastive learning on NTMs, with the aim of uncovering improved latent topic structures in documents.

\section{Method}
In this section, we will illustrate the detail of our proposed GCTM and start with the formalization used in this paper.

\subsection{Formalization}
We first introduce the overall framework of NTMs with contrastive learning.
Formally, we denote a corpus with $N$ documents as $\mathcal{D}$, where each document $d$ is represented as a sequence of words $\{w_1, w_2, \cdots, w_{n_d}\}$, and the vocabulary $\mathcal{V}$ has a size of $v$. 
For topic modelling, we assume $\theta_d$ is the document representation of the document $d$ in the latent topic space.
Global topics for the corpus are represented as $\beta$, in which each topic is a distribution over the vocabulary.
We also assume the number of topics is $k$, which is a hyperparameter.
The latent topic representation $\theta_d$ of document $d$ is assumed to be sampled from a prior distribution $p(\theta_d)$.
Due to the intractability of posterior distribution $p(\theta|x)$, NTMs use a variational distribution $q_\Theta(\theta|x)$ parameterized by an inference network with parameter set $\Theta$ to approximate it. 
Following the previous methods~\cite{iclr/SrivastavaS17-ProdLDA,icml/MiaoYB16}, we consider $\theta_d$ is sampled from a logistic normal distribution.
Based on the input feature $x_d$ of document $d$, we have:
\begin{equation}
\begin{split}
\mu_d &= f_{\mu}^d(x_d),\ \sigma_d^2 = diag(f_{\sigma}^d(x_d)), \\
\theta_d &= softmax(\mu_d + \sigma_d \epsilon_d),
\label{eq:8}
\end{split}
\end{equation}
where $f$ is the feed-forward neural network, and $\epsilon_d \in \mathcal{N}(0, I)$ is the sampled noise variable.
Then the decoder network with parameter set $\Phi$ is used to reconstruct the input $x_d$ based on $\theta_d$ and topics $\beta$. 
The objective of NTM is to maximize the evidence lower bound (ELBO) of marginal log-likelihood:
\begin{equation}
\begin{split}
    \mathcal{L}_{\mathrm{NTM}} &= \mathbb{KL}\left[q_{\Theta}(\mathbf{\theta} | x) \| p(\mathbf{\theta})\right] \\
    &-\mathbb{E}_{q_{\Theta}(\mathbf{\theta} | x)}[\log p_{\Phi}(x | \mathbf{\theta},\beta)].
\end{split}
\label{eq:9}
\end{equation}

Based on the above, contrastive learning is introduced to capture a better latent topic structure, where each document representation $\theta_d$ is associated with a positive sample $\theta^+_d$ and a negative sample $\theta^-_d$~\cite{NEURIPS2021_CLNTM}. 
The topic representations of positive pairs $(\theta, \theta^+)$ are encouraged to stay close to each other, while that of the negative pairs $(\theta, \theta^-)$ be far away from each other:
\begin{equation}
\begin{split}
    \mathcal{L}_{\text{CL}}= -\frac{1}{N} \sum_{d \in \mathcal{D}}\log \frac{\exp(\theta_{d} \cdot \theta^{+}_{d})}{\exp(\theta_{d} \cdot \theta^{+}_{d})+\alpha \cdot \exp(\theta_{d} \cdot \theta_{d}^{-})}
\end{split}
\label{eq:10}
\end{equation}
where $\alpha$ is a factor controlling the impact of negative samples.
The final optimization objective is:
\begin{equation}
    \mathcal{L} = \mathcal{L}_{\mathrm{NTM}} + \gamma
    \mathcal{L}_{\text{CL}},
\label{eq:11}    
\end{equation}
where $\gamma$ is a parameter controlling the impact of contrastive loss.
By optimizing the ELBO of NTM and the contrastive learning loss that guides the model to differentiate between positive and negative samples of input documents, it is expected that the model will uncover a superior topic structure and document representation.
Therefore, finding informative negative and positive samples is vital for efficient contrastive learning of NTMs.

\subsection{Graph Contrastive Topic Modeling}
To tackle the sample bias problem of existing methods, we aim to explore the effective sampling strategy in NTMs to generate instructive negative samples and positive samples. 
We propose the new assumption as the guidance and thus design the graph-based sampling strategy comprehensively leveraging the correlations and irrelevances among documents and words.
Based on this, we propose the graph contrastive topic model, that incorporates graph contrastive learning (GCL) into NTMs.
As shown in Figure \ref{fig:my_label}, GCTM models the input document as the document-word bipartite graph (DWBG) with the document word feature: TF-IDF.
The positive $\mathcal{W}^+$ and negative $\mathcal{W}^-$ word correlation graphs (WCGs) are built and encoded by GCNs to capture the multi-order correlations and irrelevance among words.
Based on DWBG and WCGs, we design the document-word information propagation to generate instructive negative and positive samples, based on the graph data augmentation on DWBG.
We use the contextual embedding from the pre-trained language models (PLMs)~\cite{kenton2019bert} to enrich the input feature of the prototype, design the graph contrastive loss to push close the topic representations of prototypes and positive samples, and pull away that of prototypes and negative samples.

\subsubsection{Sampling Assumption}
Existing methods~\cite{JMLR:v22:21-0089,NEURIPS2021_CLNTM,wu2022mitigating} assume that the ideal negative sample should exclude the high-frequency words of the input document.
However, it has been found that the topic of a document may not be determined by the high-frequency words but by other salient words~\cite{griffiths2019probabilistic,lau2014machine,chi2019topic}.
Therefore, there arises the question: what constitutes high-quality negative samples? 
Ideally, informative negative and positive samples should be semantically distinct and similar to the prototypes respectively~\cite{li2022uctopic}.
To answer the above question, we aim to generate the negative/positive samples that are topically unrelated/correlated with the prototypes and assume:
\begin{myDef}{two documents have distinct topics when they feature a significant disparity in their semantic content, characterized by the presence of words with dissimilar meanings.}
\end{myDef}
The common way to determine if two documents have different topics is to analyze and compare their contents characterized by the word distributions, co-occurrence patterns, and the overall context.
It is intuitive to take the semantic dissimilarity between words in two documents as the indicator to identify if they have different topics.
Based on the above hypothesis, we further assume:
\begin{myDef}{the most beneficial negative samples should encompass as many distinct words as feasible that are semantically uncorrelated with the prototype}.
\end{myDef}
We believe negative samples should include distinct words that are semantically uncorrelated with prototypes, to ensure they have different topics.
Thus, it is vital to identify words that are semantically related or irrelevant to prototypes to make efficient data augmentation.
To achieve this, we design the graph-based sampling strategy to sample desired negative and positive samples, which captures the multi-hop correlations and irrelevances between documents and words to sample desired negative and positive samples.
We will introduce it in detail in the next subsections.

\subsubsection{Graph Construction}
To fully capture semantic correlation among documents and words, we first build the input document as the document-word bipartite graph (DWBG), and the negative and positive word co-occurrence graphs (WCGs).

\textbf{Document-word Bipartite Graph}: 
DWBG captures the document-word semantic correlations.
We represent each input document $d$ with its TF-IDF feature $x_d = \{x_{d,1}, x_{d,2}, \cdots, x_{d,v}\}$.
We take $d$ as the document-word bipartite graph (DWBG) represented by the following adjacency matrix $A_d$~\cite{www/yang20-gaton,www/XieHDPN21}:
\begin{equation}
A_d = 
\begin{cases}  x_{d,i}, & \text{if $w_i$ appears in document $d$}\\
0, & \text{otherwise}
\end{cases}
\label{eq:0}
\end{equation}
For each document $d$, its DWBG contains two types of nodes: the document and all of its words, and there are only edges between the document and its words, which correspond to their respective TF-IDF values.
We further use the external knowledge from pre-trained language models~(e.g., \citealp{DBLP:conf/naacl/DevlinCLT19}) to enrich the document feature with the sequential and syntactic dependency information among words that can not be utilized by BoW features.
Following the previous method~\cite{bianchi-etal-2021-CombinedTM}, we introduce the contextual document embedding from SBERT~\cite{DBLP:conf/emnlp/ReimersG19}, which is transformed into the hidden representation with $v$-dimension via a feed-forward layer\footnote{We use the same pre-trained language model as CombinedTM~\cite{bianchi-etal-2021-CombinedTM}, i.e., \emph{stsb-roberta-large}.}.
Then, the hidden representation is concatenated with the TF-IDF feature to provide the enhanced input feature $\hat{x}_d \in \mathbb{R}^{2 \times v}$.

\textbf{Word Co-occurrence Graphs}:
We create two WCGs represented by $\mathcal{W}^+$ (positive word graph) and $\mathcal{W}^-$ (negative word graph) to save the global semantic association and irrelevance among words. 
For word co-occurrence, we employ the normalized pointwise mutual information (NPMI, ~\citealp{church-hanks-1990-npmi}). 
Formally, a word pair $(w_i, w_j)$ is denoted as:
\begin{equation}
    \mbox{NPMI}(w_i, w_j) = \frac{\log \frac{p(w_i, w_j)}{p(w_i)\cdot p(w_j)}}{-\log p(w_i, w_j)},
\label{eq:1}
\end{equation}
where $p(w_i,w_j)$ represents the probability that words $w_i$ and $w_j$ co-occur in the sliding windows, $p(w_i)$ and $p(w_j)$ refers to, respectively, the probability that words $w_i$ and $w_j$ appear in the sliding windows.

Based on it, positive and negative word graphs can be constructed.
The adjacency matrix $A^+$ of positive word graph $\mathcal{W}^+$ is denoted as:
\begin{equation}
    A^+_{ij} = 
    \begin{cases}
        \mbox{NPMI}_{ij}, & \mbox{if } i \neq j \mbox{ and NPMI}_{ij} \geq \mu^+ \\
        1, & \mbox{if } i = j \\
        0, & \text{otherwise} \\
    \end{cases}
\label{eq:2}
\end{equation}
where $\mbox{NPMI}_{ij}=\mbox{NPMI}(w_i, w_j)$ and $\mu^+$ is a non-negative threshold.
Similarly, the adjacency matrix $A^{-}$ of negative word graph $\mathcal{W}^-$ is denoted as:
\begin{equation}
    A^-_{ij} = \begin{cases}
        |\mbox{NPMI}_{ij}|, & \mbox{if } i \neq j \mbox{ and NPMI}_{ij} \leq -\mu^- \\
        0, & \text{otherwise} \\
    \end{cases}
\label{eq:3}
\end{equation}
where $|\cdot|$ is the absolute value function and $\mu^-$ is a non-negative threshold.
In contrast to prior methods~\cite{zhu-etal-2018-GraphBTM} that only considered positive word co-occurrence information, we use both positive and negative word graphs to preserve the global correlation and irrelevance among words. 
Notice that the negative word graph has no self-loops since the word is always related to itself. 

\subsubsection{Sampling Strategy}
Based on DWBG and WCGs, we formulate the data augmentation of documents as the graph augmentation problem, to generate positive and negative samples by identifying words that are semantically correlated and uncorrelated with the prototype.
We propose to use the graph convolutional network (GCN)~\cite{iclr/Kipf2017-GCN} to encode both positive and negative WCGs, to capture multi-hop correlations and the irrelevance of words.
Formally, for a one-layer GCN, node features can be calculated as
\begin{equation}
    H^{(l+1)} = \rho(\tilde {A} H^{(l)} W^{(l)}),
\label{eq:gcn}
\end{equation}
where $\rho$ is the ReLU activation function, $W^{(l)}$ is the weight matrix, $\tilde{A} = D^{-\frac{1}{2}} A D^{-\frac{1}{2}}$ is the normalized symmetrical adjacent matrix of the input graph $A$, and $D$ is the degree matrix of $A$. Given adjacency matrix $A^+$ and $A^-$, we stack $L$ layers of GCN to obtain positive/negative hidden representations of words by equation (\ref{eq:gcn}) respectively: 
\begin{equation}
\begin{split}
    \beta^+_v  = softmax\left(H_+^{(L)}\right),\quad \beta^-_v = softmax\left(H_-^{(L)}\right).
\end{split}
\label{eq:4}
\end{equation}
The input features of GCN are set to identity matrix $I$ for both settings, i.e., $H^{(0)} = I$. 
The $L^{th}$-layer output $H^{(L)} \in \mathbb{R}^{v \times k}$. 
The information propagation with $L$ GCN layers can capture the $L$-order semantic correlations/irrelevances among words in positive and negative word graphs respectively. Therefore, the hidden representation of each word is informed by its direct co-occurred/unrelated words as well as high-order neighbours.

We then design the document-word information propagation (DWIP) process, that conducts the data augmentation on the DWBG via propagating semantic correlations/irrelevances between documents and words:
\begin{equation}
    H^+_d = A_d * \beta^+_v, \quad H^-_d = A_d * \beta^-_v.
\label{eq:5}
\end{equation}
where $H^+$ and negative topic distribution $H^-$ are hidden representations of positive samples and negative samples.
The positive document hidden representation gathers information from words that are directly and distantly correlated with words that appeared in the prototype DWBG.
both words of the document and implicitly correlated words from other documents.
The negative document hidden representation is informed with words that are extremely unrelated to the prototype DWBG.
If we remove the activation function in equations (\ref{eq:gcn}) and (\ref{eq:4}), the GCN layer will degrade into the simplified GCN~\cite{DBLP:conf/icml/WuSZFYW19}, which yields:
\begin{equation}
\begin{aligned}
H^+_d &= (A_d * \tilde {A}^+) H^{(L-1)}_+ W^{(L-1)}_+,\\
H^-_d &= (A_d * \tilde {A}^-) H^{(L-1)}_- W^{(L-1)}_-.
\end{aligned}
\label{eq:6}
\end{equation}
From the perspective of graph augmentation, $x_d^+ = A_d^+ = A_d * \tilde {A}^+$ and $x_d^- = A_d^- = A_d * \tilde {A}^-$ can be interpreted as the edge perturbation-based augmentations on the input graph $A_d$. 
The edge between each document and word pair $(d,i)$ is:
\begin{equation}
\begin{aligned}
    A_{d,i}^{+} &= A_{d,i} \tilde{A}^{+}_{i,i}+\sum_{j \in \mathcal{N}_i^{+}, j \neq i} A_{d,j}\tilde{A}^{+}_{j,i}, \\
    A_{d,i}^{-} &= \sum_{j \in \mathcal{N}_i^{-}, j \neq i} A_{d,j}\tilde {A}^{-}_{j,i},
\end{aligned}
\label{eq:7}
\end{equation}
where $\mathcal{N}_i^{-},\mathcal{N}_i^{+}$ are the neighbor sets of the word $i$ in the negative and positive word graphs.

If there exists an edge between $(d, i)$ in the original graph $A_d$, which means the word $i$ is mentioned in the document $d$, the corresponding edge in $A_d^+$ will be reinforced by the neighbour words of $i$ that is also mentioned in the document $d$ with the weight $A_{d,j}$.
Otherwise, a new edge $(d,i)$ will be added to $A_d^+$ which represents the implicit correlation between word $i$ and document $d$ if $i$ is the neighbour of words that are mentioned in document $d$.
Notice that there would be no edge between word $i$ and document $d$ if word $i$ is not the neighbour of any word mentioned in the document $d$.
Thus, the sampling process is able to identify new words that latent correlate to the input document to generate positive samples.

Similarly, in $A_d^-$, the edge $A_{d,i}^-$ between word $i$ and document $d$ will be yielded if $i$ is the “fake” neighbours of any word $j$ appeared in a document $d$.
Otherwise, there will be no edge between $(d,i)$ in $A_d^-$.
Therefore, the negative samples are generated by effectively recognizing the irrelevance between words and the prototype $d$. 

\subsubsection{Contrastive Loss}
We then fed the enriched feature $\hat{x}_d$ of the prototype, the hidden representations of its negative $H^+_d$ and positive sample $H^-_d$, into the encoder to sample the latent their topic representations $(\theta_d, \theta_d^+, \theta_d^-)$ correspondingly.
The contrastive loss is calculated as in Equation \ref{eq:10} based on $(\theta_d, \theta_d^+, \theta_d^-)$, where $(\theta, \theta^+)$ are encouraged to stay close to each other, while $(\theta, \theta^-)$ be far away from each other in the latent topic space.
Finally, the model will optimize the objective with the ELBO of NTMs and the contrastive loss as in Equation \ref{eq:11}

\subsection{Understanding Graph Contrastive Topic Learning}
According to previous methods~\cite{Oord/abs-1807-03748,DBLP:conf/icml/0001I20,DBLP:conf/nips/LiPSG21,DBLP:journals/corr/abs-2107-02495}, the contrastive loss in equation (\ref{eq:10}) can be rewritten as:
\begin{equation}
    \mathcal{L}_{\text{CL}}=
    \mathbb{E}_{q(\theta,\theta')}[\log \frac{q(\theta'|\theta)}{q(\theta')}],
    \label{eq:12}
\end{equation}
where $q(\theta,\theta')$ indicates the distribution parameterized by the neural networks 
$x_d, x'_d$ can be deemed as two different views of input document $d$, and equation (\ref{eq:12}) is to maximize the mutual information of their latent representations.
Since $q(\theta'|\theta)=\frac{q(\theta,\theta')}{q(\theta)}$, equation (\ref{eq:12}) can be further rewritten as:
\begin{equation}
    \mathcal{L}_{\text{CL}}=\mathbb{E}_{q(\theta,\theta')}[\log \frac{q(\theta, \theta')}{q(\theta)q(\theta')}].
    \label{eq:13}
\end{equation}

Considering a variational auto-encoder (VAE) with two variables $x,x'$, we have the marginal likelihood:
\begin{equation}
\begin{split}
    p(x,x',\theta,\theta') &= p(x|\theta)p(x'|\theta')p(\theta,\theta'), \\
    \log p(x,x') &= \log \mathbb{E}_{q(\theta,\theta'|x,x')}[\frac{p(x|\theta)p(x'|\theta')}{q(
    \theta|x)q(\theta'|x')}p(\theta,\theta')],
\end{split}
\label{eq:14}
\end{equation}
where $q(\theta,\theta'|x,x')$ is the approximate posterior, which is usually parameterized by the encoder in VAE. 
Applying $p(x|\theta)=\frac{q(\theta|x)p_{\text{true}}(x)}{q(\theta)}$ to equation (\ref{eq:14}), we have:
\begin{equation}
\begin{split}
    \log p(x,x')&=\log \mathbb{E}_{q(\theta,\theta'|x,x')}[\frac{p(\theta,\theta')}{q(
   \theta)q(\theta')}] + const(x,x'), \\
    \log p(x,x')&=\log [\frac{p(\theta,\theta')}{q(
   \theta)q(\theta')}] + const(x,x').
    \end{split}
    \label{eq:15}
\end{equation}
Notice that $p_{\text{true}}(x), p_{\text{true}}(x')$ are parameter-free and constants. 
Once we have the deterministic encoder to approximate the posterior $q(\theta,\theta'|x,x')$, it can be collapsed from equation (\ref{eq:15}).
Then we further rewrite equation (\ref{eq:15}) by introducing $q(\theta, \theta')$ similar to \citet{DBLP:journals/corr/abs-2107-02495}:
\begin{equation}
\begin{split}
    &\log p(x,x') = \log \mathbb{E}_{p_{\text{true}}(x,x')}[\frac{p(\theta,\theta')}{q(
    \theta)q(\theta')}] + const\\
    &=\log \mathbb{E}_{q(\theta, \theta')} [\frac{q(\theta,\theta')}{q(
    \theta)q(\theta')}]+\log [\frac{p(\theta,\theta')}{q(\theta,\theta')}] + const.
\end{split}
\label{eq:16}
\end{equation}
We can see that the first term is actually the same as the contrastive loss in equation (\ref{eq:13}).
If we let $p(\theta,\theta')=q(\theta,\theta')$, the second term is zero, and equation (\ref{eq:16}) is totally the same as the contrastive loss. Interestingly, we find that the contrastive loss can be interpreted as the structured variational graph auto-encoder with two latent variables on the input document graph $x_d$ and its augmentation $x'_d$. Existing NTMs are actually variational auto-encoders with one latent variable, which aim to learn a good function $q(\theta|x)$ and force it close to the prior $p(\theta)$ in the meantime, while the contrastive loss aims to learn a mapping $q(\theta,\theta')$ and push it close to the prior $p(\theta,\theta')$. Obviously, the augmentation $\theta'$ provides extra information to better model the real data distribution of documents. This makes $q(\theta,\theta')$ better capture the latent topic structure.

\section{Experiments}
\subsection{Datasets and Baselines}
To evaluate the effectiveness of our method, we conduct experiments on three public benchmark datasets, including 20 Newsgroups (20NG) \footnote{\url{http://qwone.com/~jason/20Newsgroups/}},
IMDB movie reviews \cite{maas-2011-IMDB}, and Neural Information Processing Systems (NIPS) papers from 1987 to 2017 \footnote{\url{https://www.kaggle.com/datasets/benhamner/nips-papers}}.
As the statistics shown in Table \ref{dataset}, these corpora in different fields have different document lengths and vary in vocabulary size.
Following previous work~\cite{NEURIPS2021_CLNTM}, we adopt the same train/validation/test split as it: 48\%/12\%/40\% for 20NG dataset, 50\%/25\%/25\% for IMDB dataset, and 60\%/20\%/20\% for NIPS dataset.
For preprocessing, we utilize the commonly-used script published by \citet{card-etal-2018-SCHOLAR} \footnote{\url{https://github.com/dallascard/scholar}}, which would tokenize and remove special tokens such as stopwords, punctuation, tokens containing numbers, and tokens with less than three characters.
The parameter settings are listed in Table \ref{tab:parameters}.

\begin{table}[ht]
\caption{Details of hyperparameters.}
    \centering
    \begin{tabular}{c|c}
        \hline
        Hyperparameters & Values \\
        \hline
        learning rate & \{0.001 $\sim$ 0.007\} \\
        epochs & \{200, 300, 400, 500\} \\
        batch size & \{200, 500\}  \\
        weight $\alpha$ & \{0.5, 1, $e^{0.5}$, 2, 3, 5\} \\
        weight $\gamma$ & \{1\} \\
        \#GCN layer $L$ & \{1, 2, 3\} \\
        \hline
    \end{tabular}
    \label{tab:parameters}
\end{table}
For baseline methods, we use their official codes and default parameters. As for CLNTM + BERT and our model, we search for the best parameters from \{lr: 0.001, 0.002\} and \{epochs: 200, 300, 400, 500\} on 20NG and IMDB datasets, \{lr: 0.002, 0.003, 0.004, 0.005, 0.006, 0.007\} and \{epochs: 300, 400, 500\} on NIPS dataset. We use batch size 200 for 20NG and IMDB, 500 for NIPS, and Adam optimizer with momentum 0.99 across all datasets. 
The weight $\alpha$ is set to $e^{0.5}$ on 20NG and IMDB datasets and 1 on the NIPS dataset, which is the best values in our setting. But one can simply set $1 \leq \alpha \leq 2$ if time is limited, which hardly hurts the performance. The weight of contrastive loss $\gamma$ is set to 1.

\begin{table}[!htb]
\caption{Statistics of the datasets.}
\centering
\normalsize
\begin{tabular}{cccc}
\hline
\textbf{Dataset} & \textbf{\#Train} & \textbf{\#Test} & \textbf{Vocab}\\
\hline
20NG & {11314} & {7532} & {2000}\\ 
IMDB & {25000} & {25000} & {5000}\\ 
NIPS & {5792} & {1449} & {10000}\\ 
\hline
\end{tabular}
\label{dataset}
\end{table}
We compare our method with the following baselines:
\begin{enumerate}
    \item NTM~\cite{icml/MiaoYB16} is a neural topic model based on the Gaussian prior.
    \item ProdLDA~\cite{iclr/SrivastavaS17-ProdLDA} is a groundbreaking work that uses the AEVB inference algorithm for topic modelling. 
    \item SCHOLAR \cite{card-etal-2018-SCHOLAR} extends NTM with various metadata. 
    \item SCHOLAR+BAT \cite{hoyle-2020-BAT+SCHOLAR} fine-tunes a pre-trained BERT autoencoder as a “teacher” to guide a topic model with distilled knowledge.
\item W-LDA~\cite{nan-etal-2019-WLDA} enables Dirichlet prior on latent topics by the Wasserstein autoencoder framework.
\item BATM~\cite{wang-etal-2020-Gaussian-BAT} introduces generative adversarial learning into a neural topic model.
\item CombinedTM \cite{bianchi-etal-2021-CombinedTM} extends ProdLDA by combining contextualized embeddings with BoW as model input.
\item CLNTM \cite{NEURIPS2021_CLNTM} introduces the contrastive learning loss for NTM. Its positive and negative samples are from substituting some items of BoW with corresponding items of reconstructed output.
\end{enumerate}
To evaluate the quality of generated topics from our method and baselines, we employ the normalized pointwise mutual information (NPMI) based on the top 10 words of each topic on all datasets.
It is reported that NPMI is an automatic evaluation metric for topic coherence which is highly correlated with human judgements of the generated topic quality \cite{nips/Chang09-reading-tea-leaves, lau-2014-machine-reading-tea-leaves}.
We report the mean and standard deviation of 5 runs with different random seeds.
\subsection{Results and Analysis}
\subsubsection{Main Results}
We first present the NPMI results of our method and baselines on three datasets.
As shown in Table \ref{results}, our method yields the best performance on all datasets with two different topic number settings.
\begin{table*}[!htb]
\footnotesize
\caption{NPMI coherence of the models on three datasets. The best scores are marked in bold. Some results are referenced from~\citet{NEURIPS2021_CLNTM,hoyle-2020-BAT+SCHOLAR}.}
\centering
\begin{tabular}{ccccc}
\hline
\multirow{2}{*}{\textbf{Model}} & \textbf{20NG} & \textbf{IMDB} & \textbf{NIPS} \\
 & $k$ = 50 \qquad \qquad $k$ = 200 & $k$ = 50 \qquad \qquad $k$ = 200 & $k$ = 50 \qquad \qquad $k$ = 200  \\
\hline
NTM & 0.283 $\pm$ 0.004 \ 0.277 $\pm$ 0.003 & 0.170 $\pm$  0.008 \ 0.169 $\pm$  0.003  & - \\
ProdLDA & 0.258 $\pm$ 0.003 \ 0.196 $\pm$ 0.001 & 0.134 $\pm$ 0.003 \ 0.112 $\pm$ 0.001 & 0.199 $\pm$ 0.006 \ 0.208 $\pm$ 0.006 \\
W-LDA & 0.279 $\pm$ 0.003 \ 0.188 $\pm$ 0.001 & 0.136 $\pm$ 0.007 \ 0.095 $\pm$0.003 & - \\
BATM& 0.314 $\pm$  0.003 \ 0.245 $\pm$  0.001 & 0.065 $\pm$ 0.008 \ 0.090 $\pm$ 0.004 & - \\
SCHOLAR  & 0.297 $\pm$ 0.008 \ 0.253 $\pm$ 0.003 & 0.178 $\pm$ 0.004 \ 0.167 $\pm$ 0.001 & 0.389 $\pm$ 0.010 \ 0.335 $\pm$ 0.002  \\
SCHOLAR+BAT & 0.353 $\pm$ 0.004 \ 0.315 $\pm$ 0.004 & 0.182 $\pm$ 0.002 \ 0.175 $\pm$ 0.003 & 0.398$\pm$ 0.012 \ 0.344 $\pm$ 0.005  \\
CombinedTM   & 0.263 $\pm$ 0.004 \ 0.209 $\pm$ 0.004 & 0.125 $\pm$ 0.006 \ 0.101 $\pm$ 0.004 & 0.201 $\pm$ 0.004 \ 0.212 $\pm$ 0.004  \\
CLNTM        & 0.327 $\pm$ 0.005 \ 0.276 $\pm$ 0.002 & 0.183 $\pm$ 0.005 \ 0.175 $\pm$ 0.002 & 0.391 $\pm$ 0.009 \ 0.337 $\pm$ 0.005  \\
CLNTM+BERT & 0.329 $\pm$ 0.003 \ 0.292 $\pm$ 0.005 & 0.196 $\pm$ 0.003 \ 0.183 $\pm$ 0.003 &  0.396 $\pm$ 0.007 \ 0.345 $\pm$ 0.002  \\
GCTM         & \textbf{0.379 $\pm$ 0.004} \ \textbf{0.338 $\pm$ 0.005}  & \textbf{0.202 $\pm$ 0.004} \ \textbf{ 0.184$\pm$0.007 } & \textbf{0.410 $\pm$ 0.006} \ \textbf{0.387 $\pm$ 0.008} \\
\hline
\end{tabular}
\label{results}
\end{table*}
It demonstrates the effectiveness of our method which exploits in-depth semantic connections between documents and words via graph contrastive learning.
Compared with CLNTM and its extension CLNTM+BERT with contextual document embeddings that also introduce contrastive learning into topic modelling, our method presents improvements on all datasets.
In their methods, only the direct semantic correlations are considered when selecting negative and positive samples for contrastive learning, resulting in noisy negatives that may be semantically related to the source document and misleading the model.
Different from their methods, our method can exploit the multi-hop interactions between documents and words based on the document-word graph along with positive and negative word correlation graphs.
It allows our method to precisely generate samples for graph contrastive learning, leading to a better understanding of the source document.

Our method also outperforms previous neural topic models enriched by pre-trained contextual representations such as SCHOLAR+\\BAT.
It proves the essential of contrastive learning for neural topic modelling, which is that it can learn more discriminative document representations.
This is also proved in CombinedTM which shows poor performance in all datasets, 
since it directly concatenates the contextualized representations with word features without contrastive learning.

\subsubsection{Parameter Sensitivity}

\paragraph{Topic number.} As shown in Table \ref{results}, we present the results of our method in two different topic number settings as $k=50$ and $k=200$.
Our method with 50 topics outperforms that with 200 topics on all datasets, which is similar to other baselines.
We guess that redundant topics would be generated when the predefined topic number is much larger than the real number of semantic topics.

\begin{figure}[ht]
    \centering
    \includegraphics[width=0.5\linewidth]{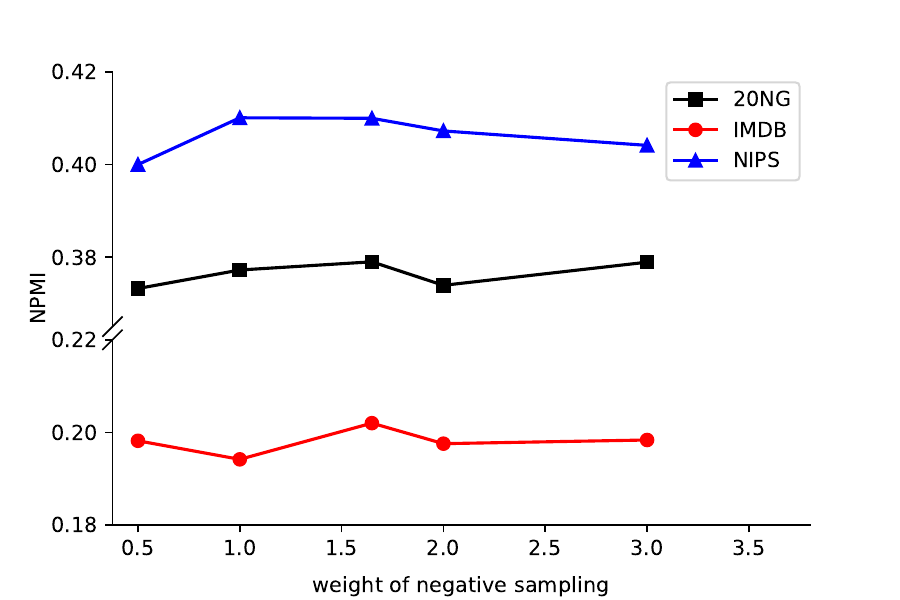}
    \caption{Results of our model with different weights of negative samples.}
    \label{weight}
\end{figure}

\paragraph{Weight of negative sampling.}
We also perform a series of experiments with different $\alpha$ in equation (\ref{eq:10}) to investigate the impact of the weight of negative samples.
As shown in Figure~\ref{weight}, the NPMI first increases with the growth of $\alpha$.
The model achieves the highest scores when $1\leq \alpha \leq 2$ and presents worse results with a larger $\alpha$.
This indicates that the weight of the negative samples should not be too large or too small since it affects the gradient of contrastive loss $\mathcal{L}_{\text {CL}}$ (equation \ref{eq:10}) with respect to latent vector $\theta_d$ \cite{NEURIPS2021_CLNTM}.

\begin{figure}[ht]
    \centering
    \includegraphics[width=0.5\columnwidth]{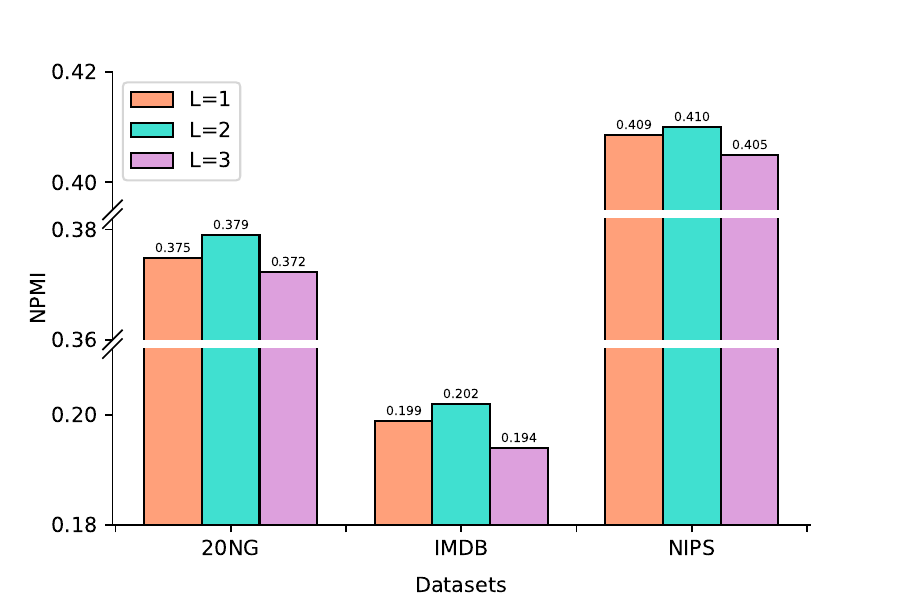}
    \caption{Effect of the number of GCN layers. The results are the average of 5 runs.}
    \label{fig:GCN_layers}
\end{figure}

\paragraph{GCN layers.}
We encode the word graphs with different numbers of GCN layers in our method to evaluate the effect, as shown in Figure~\ref{fig:GCN_layers}.
On all three datasets, the model performs better with two GCN layers than one layer, but the performance drops dramatically when $L$ increases to 3.
Similar to \citet{iclr/Kipf2017-GCN}, 
we argue that stacking too many GCN layers (e.g., more than 2) could introduce extra noise due to excessive message passing, while one GCN layer can only exploit limited structural information of the graphs. 

\begin{table*}[!htb]
\caption{Comparison of topics (top 10 words) from different models on NIPS dataset. The topic labels are annotated manually. The number of topics is 50.}
\centering
\small
\begin{tabular}{cccc}
\hline
\textbf{Model} & \textbf{NPMI} & \textbf{Topics} \\
\hline
\multirow{4}{*}{SCHOLAR+BAT} & \multirow{2}{*}{0.6460} & mead silicon chip fabrication neuromorphic & \multirow{2}{*}{chip}\\
                               &                         & photoreceptors fabricated cmos mosis chips &  \\
                               & \multirow{2}{*}{0.5534} &  bellman policy tsitsiklis mdps mdp        & \multirow{2}{*}{RL}\\
                               &                         & parr discounted policies ghavamzadeh approximator \\
\multirow{4}{*}{CLNTM}  & \multirow{2}{*}{0.6575} & vlsi chip cmos analog chips & \multirow{2}{*}{chip}\\
                        &                         & fabricated programmable transistors mosis circuit  \\
                        & \multirow{2}{*}{0.5491} & reinforcement bellman tsitsiklis mdp mdps & \multirow{2}{*}{RL}\\
                        &                         & athena bertsekas discounted ghavamzadeh policy \\
\multirow{4}{*}{Ours}   & \multirow{2}{*}{\textbf{0.6728}} & cmos transistors fabricated transistor chip & \multirow{2}{*}{chip}\\
                        &                         & chips voltages analog programmable capacitor \\
                        & \multirow{2}{*}{\textbf{0.6158}} & mdps mdp policy bellman discounted & \multirow{2}{*}{RL}\\
                        &                         & policies horizon rewards reinforcement parr \\
\hline
\end{tabular}
\label{topics-nips}
\end{table*}
\begin{table*}[ht]
\caption{Comparison of topics from different models on 20NG dataset.}
\centering
\small
\begin{tabular}{cccc}
\hline
\textbf{Model} & \textbf{NPMI} & \textbf{Topics} \\
\hline
\multirow{4}{*}{SCHOLAR+BAT} & \multirow{2}{*}{0.5654} & encryption enforcement wiretap escrow clipper & \multirow{2}{*}{security}\\
                               &                         & secure encrypt cryptography security agency &  \\
                               & \multirow{2}{*}{0.3778} &  moon pat orbit flight earth        & \multirow{2}{*}{space flight}\\
                               &                         & lunar fly fuel nasa space \\
\multirow{4}{*}{CLNTM}  & \multirow{2}{*}{0.5691} & clipper escrow enforcement secure keys & \multirow{2}{*}{security}\\
                        &                         & wiretap agencies chip algorithm encryption  \\
                        & \multirow{2}{*}{0.3552} & orbit space shuttle mission nasa & \multirow{2}{*}{space flight}\\
                        &                         & vehicle earth station flight ron \\
\multirow{4}{*}{Ours}   & \multirow{2}{*}{\textbf{0.5941}} & encryption escrow clipper secure crypto & \multirow{2}{*}{security}\\
                        &                         & nsa keys key wiretap privacy \\
                        & \multirow{2}{*}{\textbf{0.4340}} & orbit space launch moon shuttle & \multirow{2}{*}{space flight}\\
                        &                         & henry nasa flight spencer earth \\
\hline
\end{tabular}
\label{topics-20ng}
\end{table*}
\subsubsection{Ablation Study}
To verify the contribution of each component in our proposed method, we adopt different objectives to train the model and evaluate the performance, including:
1) w/o cl: with only NTM loss; 2) w/o neg: with only positive samples; 3) w/o pos: with only negative samples; 4) full: with contrastive loss and NTM loss. 
The corresponding loss functions are as followed:\\
1) without contrastive loss:
\begin{equation*}
    \mathcal{L} = \mathcal{L}_{\mathrm{NTM}};
\end{equation*}
2) without negative sampling:
\begin{equation*}
    \mathcal{L} = \mathcal{L}_{\mathrm{NTM}} -\frac{\gamma}{N}\sum_{d \in \mathcal{D}}  \theta_d \cdot \theta^+_d;
\end{equation*}
3) without positive sampling:
\begin{equation*}
    \mathcal{L} = \mathcal{L}_{\mathrm{NTM}} +\frac{\gamma}{N}\sum_{d \in \mathcal{D}}  \theta_d \cdot \theta^-_d;
\end{equation*}
4) with full contrastive loss:
\begin{equation*}
    \mathcal{L} = \mathcal{L}_{\mathrm{NTM}} + \gamma \mathcal{L}_{\text{CL}}.
\end{equation*}

The results are shown in Table \ref{ablation study}. 
Among all ablations, w/o cl presents the lowest NPMI scores, which proves again the essential of contrastive learning in topic modelling.
It can also be observed that the performance of our method compromises without either positive or negative samples, demonstrating that both positive and negative samples can improve the quality of generated topics.
The decrease in NPMI scores for our method without positive samples is more significant than that without negative samples.
Similar to previous work~\cite{NEURIPS2021_CLNTM}, the improvement of our method can attribute more to positive samples than negative samples.
Moreover, positive and negative samples are complementary in generating coherent topics, resulting in the best scores with full contrastive loss.
\begin{table}[ht]
\caption{Ablation study ($k$ = 50).}
\centering
\normalsize
\begin{tabular}{ccccc}
\hline
 & \textbf{20NG} & \textbf{IMDB} & \textbf{NIPS} \\
\hline
w/o cl    & 0.362 $\pm$ 0.008 & 0.193 $\pm$ 0.006 & 0.401 $\pm$ 0.004  \\
w/o neg   & 0.373 $\pm$ 0.008 & 0.198 $\pm$ 0.003 & 0.406 $\pm$ 0.008  \\
w/o pos  & 0.369 $\pm$ 0.005 & 0.195 $\pm$ 0.005 & 0.404 $\pm$ 0.007  \\
Full& \textbf{0.379 $\pm$ 0.005}  & \textbf{0.202 $\pm$ 0.004} & \textbf{0.410 $\pm$ 0.006}\\
\hline
\end{tabular}
\label{ablation study}
\end{table}

\subsubsection{Case Study}
We randomly sample several topics from different models on NIPS and 20NG datasets to investigate the quality of our generated topics in Table~\ref{topics-nips} and Table \ref{topics-20ng}, respectively.
It clearly shows that our model yields more coherent and interpretable topics than baselines. For example, in the two selected topics that can be described as "chip'' and “reinforcement learning'', the word “mead” extracted from SCHOLAR + BAT and "ghavamzade" from CLNTM are not quite consistent with the topics. In contrast, almost all words extracted from our model are in line with the related topics.
\subsubsection{Text Classification}
In order to evaluate the representation capability of our generated document representations, we resort to downstream application performance, i.e., text classification accuracy.
Following previous methods, we utilize the generated document representations of our method with 50 topics to train a Random Forest classifier.
As shown in Table~\ref{classification}, our method presents the best classification results, which demonstrates the benefit of our meaningful representations on predictive tasks.
\begin{table}[ht]
\caption{Classification accuracy of different models on 20NG and IMDB datasets (comprising of 20 and 2 classes respectively). The topic number $k = 50$.}
\normalsize
    \centering
    \begin{tabular}{cccc}
        \hline
        Model         & 20NG  & IMDB \\
        \hline
        SCHOLAR       & 0.452 & 0.855 \\
        SCHOLAR + BAT & 0.421 & 0.833 \\
        CLNTM         & 0.486 & 0.827 \\
        CLNTM + BERT  & 0.525 & 0.871 \\
        Ours          & \textbf{0.543} & \textbf{0.874} \\
        \hline
    \end{tabular}
    \label{classification}
\end{table}

\section{Conclusion}
In this paper, we propose a novel graph-based sampling strategy and propose a novel graph contrastive neural topic model incorporating graph-contrastive learning for document topic modeling.
Experimental results prove the superiority of our proposed method due to the better sampling strategy based on graph augmentation with multi-order semantic correlation/irrelevance among documents and words.
We show that the contrastive learning in our model is actually a structured variational auto-encoder, thus it can better model the data distribution of documents to learn better latent topics.
Since the NTM loss is a variational auto-encoder with a single latent variable, we argue that the contrastive loss can actually cover the learning process in NTM loss, but more effort is required to explore and verify it.
In the future, we will explore removing the traditional NTM loss and further investigate the effectiveness of pure contrastive loss on document topic modeling.




\bibliographystyle{ACM-Reference-Format}
\bibliography{sample-base}

\appendix

\end{document}